\documentclass[runningheads]{llncs}
\usepackage{graphicx}
\usepackage{dirtytalk}
\usepackage{tabularx}
\usepackage{amsmath}
\usepackage{multirow}

\usepackage[dvipsnames]{xcolor}
\usepackage{pgfplots}
\pgfplotsset{width=30cm,compat=1.9}
\usetikzlibrary{patterns}

\begin{document}
\title{Prompt Agnostic Essay Scorer: A Domain Generalization Approach to Cross-prompt Automated Essay Scoring}
\titlerunning{Prompt Agnostic Essay Scorer}
%
\author{Robert Ridley \and
Liang He \and
Xinyu Dai \and
Shujian Huang \and
Jiajun Chen}

\institute{National Key Laboratory for Novel Software Technology Nanjing University, Nanjing 210023, China \\
\email{robertr@smail.nju.edu.cn}}
\maketitle              
\begin{abstract}
Cross-prompt automated essay scoring (AES) requires the system to use non target-prompt essays to award scores to a target-prompt essay. Since obtaining a large quantity of pre-graded essays to a particular prompt is often difficult and unrealistic, the task of cross-prompt AES is vital for the development of real-world AES systems, yet it remains an under-explored area of research. Models designed for prompt-specific AES rely heavily on prompt-specific knowledge and perform poorly in the cross-prompt setting, whereas current approaches to cross-prompt AES either require a certain quantity of labelled target-prompt essays or require a large quantity of unlabelled target-prompt essays to perform transfer learning in a multi-step manner. To address these issues, we introduce Prompt Agnostic Essay Scorer (PAES) for cross-prompt AES. Our method requires no access to labelled or unlabelled target-prompt data during training and is a single-stage approach. PAES is easy to apply in practice and achieves state-of-the-art performance on the Automated Student Assessment Prize (ASAP) dataset.

\end{abstract}
\section{Introduction} \label{introduction}
Automated Essay Scoring (AES) is the task of using computation to assign a score to a piece of writing. An effective AES system would bring about numerous benefits to the field of education, including saving teachers time spent grading papers, removing teacher bias towards students, and providing students with instant feedback on their written work. \par
The majority of research into AES focuses on prompt-specific essay scoring, whereby a model is created for each specific essay prompt in the corpus \cite{williamson}. While many recent approaches to prompt-specific essay scoring have achieved a high level of performance \cite{alikaniotis,taghipour,dong1,dong2,tay}, they are ineffective when applied in a cross-prompt setting, whereby there is no access to labelled target-prompt essays, only to essays from other prompts \cite{jin}. Their ineffectiveness arises due to an over-reliance on prompt-specific knowledge. Therefore, when there are differences between the training data and test data distributions, these models overfit to the training data and exhibit a significant drop in performance. Since, in practice, it is common to have an insufficient quantity of labelled essays for the target prompt, effective cross-prompt AES systems are vital for real-world applications. \par
To address the issue of insufficient quantity of labelled target-prompt essays, some transfer learning approaches have been proposed. Phandi, et al. \cite{phandi} apply a correlated Bayesian linear ridge regression algorithm to adapt their AES system from an initial prompt to a new prompt, and Cummins et al. \cite{cummins} treat the problem as a multi-task learning problem and introduce a constrained preference-ranking approach. While these systems are less reliant on prompt-specific knowledge, they still require a certain number of labelled target-prompt essays to achieve acceptable levels of performance. \par
In order to tackle the problem of cross-prompt AES in the scenario where there are no labelled target-prompt essays available for training, Jin et al. \cite{jin} propose assigning pseudo labels to target-prompt essays as part of a two-stage approach. This method overcomes the need for labelled target-prompt essays. However, it still requires a high quantity of unlabelled target-prompt essays to which to assign pseudo labels and it also relies on there being a good distribution of high- and low-quality target-prompt essays. \par
In consideration for the difficulties of cross-prompt AES and the aforementioned issues, we propose Prompt Agnostic Essay Scorer (PAES)\footnote{Our code will be available on Github}, a neural network that utilizes non prompt-specific, general features to score essays for cross-prompt AES where there are no target-prompt essays available for training. The contributions of this paper are as follows:
\begin{itemize}
    \item To the best of our knowledge, we are the first to apply a neural-network architecture combined with traditional linguistic features in a single-stage approach to the task of cross-prompt AES, avoiding the need for pseudo-labelling, the need for abundant unlabelled target-prompt essays, and the need for a suitable distribution of quality in the target-prompt essays.
    \item Our novel approach to cross-prompt AES combines syntactic representations and non prompt-specific features to represent essay quality, enabling the model to avoid overfitting to the non target-prompt training data, an issue that is prevalent in current state-of-the-art prompt-specific approaches.
    \item Through extensive experiments, we demonstrate that our approach achieves state-of-the-art performance in the task of cross-prompt AES.
\end{itemize}
The remainder of this paper is organised as follows: In section \ref{related}, we introduce the existing literature on AES. Then, in section \ref{approach}, we describe our approach in detail. After that, we detail our experiments and provide a thorough analysis of our results in section \ref{experiments}. Finally, in section \ref{conclusions} we outline our conclusions.

\section{Related Work} \label{related}
\subsection{Prompt Specific AES}
Considerable research has gone into exploring the use of handcrafted features combined with traditional machine learning algorithms. Length-based, lexical, grammatical, discourse, cohesion, readability and semantic features have been thoroughly researched \cite{chen,larkey,rudner,attali,mcnamara,lei,smith} and have been applied to either regression \cite{page,attali}, classification \cite{rudner} or ranking \cite{chen,yannakoudakis} algorithms. \par
More recently, neural-network-based methods have brought success in prompt-specific AES, with many approaches achieving results comparable to human raters. Taghipour and Ng \cite{taghipour} apply convolutional and recurrent layers on top of one-hot encoded essay sequences; Alikaniotis et al. \cite{alikaniotis} train score-specific word embeddings; Dong and Zhang \cite{dong1} use a hierarchical model to learn sentence-level and essay-level representations; and Dong et al. \cite{dong2} utilize attention pooling \cite{sutskever}. \par
These neural-based approaches utilize word embeddings to represent the essay sequences. This enables the effective learning of semantic representations, giving a performance boost in the task of prompt-specific AES. This is, however, a big drawback in cross-prompt AES, where the semantic space of the training data can differ vastly from that of the test data, leading these models to overfit to the training data and experience drops in performance.

\subsection{Cross-prompt AES}
Cross-prompt AES requires training a model on non target-prompt essays in order to score essays from a target prompt. \par
Some methods -- including Phandi et al. \cite{phandi}, who apply correlated Bayesian linear ridge regression, and Cummins et al. \cite{cummins}, who perform multi-task learning -- apply transfer learning while assuming a small quantity of labelled target-prompt essays are available for training. However, while these methods attempt to utilize non-target prompt essays to overcome the issue of insufficient labelled target-prompt data, they still require enough labelled target-prompt essays to be effective. \par
Jin et al. \cite{jin} assume that no labelled target-prompt essays are available for training and propose a two-stage approach named \textit{TDNN}. In the first stage, they assign pseudo labels \{0,1\} to the lowest and highest quality target-prompt essays, which they do through extracting prompt-independent features and training a supervised model on the non target-prompt essays. The assumption is that, no matter which prompt the essays belong to, those of lowest and highest quality should be identifiable with prompt-independent features, such as the number of typos, grammar errors, etc.. The pseudo-labelled target-prompt essays are then used as training data for a neural network with prompt-specific features to make predictions on all target-prompt essays. Even though this method can achieve good performance when there are no labelled target-prompt essays, it still requires a high quantity of unlabelled target-prompt essays to which to assign pseudo labels. When the number of unlabelled target-prompt essays is insufficient, the neural network in the second stage will not have enough data to train on, leading to a significant performance drop. This model also relies on there being a good distribution of high- and low-quality target-prompt essays. If there are very few target-prompt essays of extreme quality, then the error for the pseudo labels will be large, leading to an overall decrease in performance. \par

\section{Approach} \label{approach}

\begin{figure}
\includegraphics[width=\textwidth]{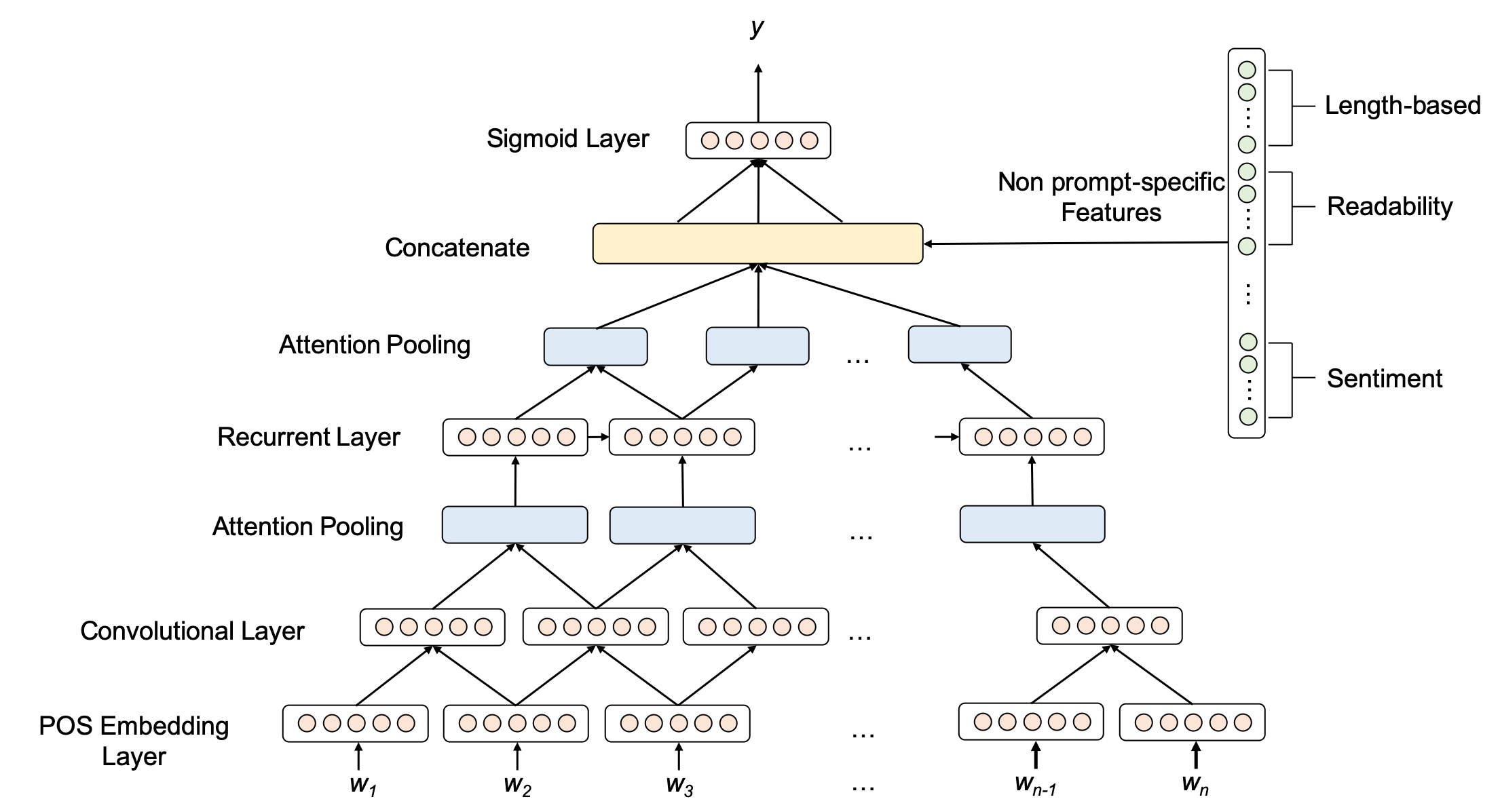}
\caption{Architecture of \textit{PAES} model} \label{fig1}
\end{figure}

Motivated by the issue of overfitting in prompt-specific approaches and the issues of requiring large quantities of unlabelled target-prompt essays and suitable distributions of essay quality, we design a single-stage neural-based approach named \textit{PAES}, depicted in Fig. \ref{fig1}. Our model is described in detail in Section \ref{model}\footnote{Refer to Appendix for objective function, optimizer and hyperparameter settings.}.

\begin{table}
\caption{Two truncated essay prompts from The Automated Student Assessment Prize (ASAP) dataset}\label{tab1}
\begin{tabularx}{\textwidth}{|X|X|}
\hline
Prompt 1 &  Prompt 2\\
\hline
 Write a letter to your local newspaper in which you state your opinion on the effects computers have on people. Persuade the readers to agree with you & Write a persuasive essay to a newspaper reflecting your views on censorship in libraries. \\
\hline
\end{tabularx}
\end{table}

\subsection{Model} \label{model}
Following the effectiveness of representing the hierarchical structure of essays in the task of prompt-specific AES, we adopt an attention-based recurrent convolutional neural network \cite{dong2} as our base model. However, instead of using word embeddings, which encourage overfitting in the cross-prompt setting, we represent the essay text through part-of-speech embeddings due to their ability to obtain better generalized representations. We also extract a set of non prompt-specific features to capture the essay quality from different aspects. These features are concatenated with the essay representation before the score is predicted through a sigmoid activation function. Our entire method is encapsulated in a single step, described in detail as follows:

\subsubsection{Essay Representation} 
One challenge of cross-prompt AES is the issue of semantic-space disparity between the essays of different prompts. Consider the essay prompts from Table \ref{tab1}. Prompt 1 requires the student to write his/her opinion about computers, and Prompt 2 requires an opinion about censorship in libraries. The sentence ``I think computers are necessary" is far more likely to appear in Prompt 1 essays than Prompt 2 essays. Whereas, ``I believe censorship is useful" has a higher likelihood of appearing in Prompt 2 essays. In terms of quality, these two sentences may be equally `good' in their own domains, but they have very different semantic representations. For this reason, directly using word embeddings as features is not suitable for cross-prompt AES. Therefore, we propose using part-of-speech embeddings to represent the essay text syntactically. This enables a more generalized representation. While the two example sentences have very different semantics, they can both be represented by the part-of-speech sequence \textit{pronoun $\rightarrow$ verb $\rightarrow$ noun $\rightarrow$ verb $\rightarrow$ adjective}. It is therefore clear that representing essay sequences with part-of-speech tags is more adaptable to different prompts than word embeddings.\par

We obtain a syntactical representation of the essay text through learning dense part-of-speech representations. The essay sequence is first tokenized and POS-tagged with the python NLTK\footnote{http://www.nltk.org} package. The sequence of POS-tagged words \(w_1,w_2,...,w_n\) is then mapped onto a dense vector \(x_i\), \(i=1,2,...,n\).
\begin{equation} \label{eq1}
x_i = \textbf{Ew}_i, i = 1,2,...,n
\end{equation}
where \(w_i\) is the one-hot representation for the POS tag of the \(i\)-th word in the sentence, \textbf{E} is the embedding matrix, and \(x_i\) is the embedding vector of the POS tag for the \(i\)-th word. \par
We then apply a 1D convolutional\footnote{Refer to Appendix for formal explanations of the modules in \textit{PAES}.} layer on top of the POS embeddings to obtain n-gram representations. This is followed by attention pooling \cite{dong2} to capture sentence-level representations. \par
We apply LSTMs \cite{hochreiter,pascanu} over the sentence vectors in the Recurrent Layer, which is then followed by another attention pooling \cite{dong2} layer that obtains an overall essay representation.

\subsubsection{Non Prompt-specific Features} \label{features} In order to represent the quality of the essays from different perspectives, we extract a variety of non prompt-specific features, which are categorized as follows:
\begin{itemize}
    \item \textit{Length-based:} Previous research \cite{chen,jin} has demonstrated the use of length-based features in AES, with longer words, sentences and essays often being indicators of higher quality of writing.
    \item \textit{Readability:} Readability scores inform on how difficult a text is to read. In well-written essays, wide vocabulary usage and varied sentence structures increase the reading difficulty of the text \cite{ke}. We utilize a variety of readability scores, which were extracted with the readability\footnote{https://pypi.org/project/readability} and textstat\footnote{https://pypi.org/project/textstat} packages.
    \item \textit{Text Complexity:} More complex texts will generally indicate a higher level of writing ability. As with Jin et al. \cite{jin}, we extract the number of clauses\footnote{Essays are parsed with spacy: https://spacy.io.} per sentence, the mean clause length, the maximum number of clauses in a sentence, the average parse tree depth per sentence in each essay and the average parse depth of each leaf node.
    \item \textit{Text Variation:} Motivated by the intuition that well-written essays will contain richer vocabulary and a variety of structures, we capture text variation through features such as unique-word count, part-of-speech tag counts, the proportion of stopwords, etc.
    \item \textit{Sentiment:} Well-written works are able to evoke emotions in the reader and are thus rich with sentiment information. We utilize the sentiment intensity analyser package from NLTK to capture the proportion of sentences that contain positive, negative and neutral sentiment.
\end{itemize}

There are a total of \(M\) feature categories \(c_1\), \(c_2\),...,\(c_M\), where \(c_i\) has a set of features \(F_i\).
A feature vector \(\textbf{f}\) containing the features \(f_1,f_2,...,f_m\) is constructed, where \(m = \sum_{i=1}^M|F_i|\). Due to variations in terms of scale and distribution of features across the different essay sets, we do set-wise feature normalization, whereby the features in each essay set are normalized independently. The features are scaled such that for feature \(f_i\) in feature vector \(\textbf{f}\), \(f_i \in [0,1]\).

\subsubsection{Concatenation} The essay representation vector \textbf{o} and non prompt-specific feature vector \textbf{f} are concatenated, giving the final essay representation \(\textbf{e}\):

\begin{equation} \label{eq15}
\textbf{e} = [\textbf{o};\textbf{f}]
\end{equation}

\subsubsection{Sigmoid Layer} We treat the task as a regression problem. A linear layer is applied to the essay representation followed by a sigmoid activation to predict the essay score.

\begin{equation} \label{eq16}
\hat{y} = \sigma(\textbf{w}_y \cdot \textbf{e} + \textbf{b}_y)
\end{equation}

where \(\hat{y}\) is the predicted score; \(\textbf{w}_y\) is a weights vector; \(\textbf{b}_y\) is a bias vector; and \(\sigma\) denotes the sigmoid activation function.

\section{Experiments} \label{experiments}
\subsection{Setup}

\begin{table}
\caption{Summary of ASAP dataset; in the Genre column, ARG represents \textit{argumentative}, RES represents \textit{response} and NAR represents \textit{narrative}.}\label{dataset}
\begin{tabular}{|l|l|l|l|l|l|l|l|}
\hline
Set & Num Essays & Genre & Mean Length & Score Range & Min score count & Max score count \\
\hline
 1 & 1783 & ARG & 350 & 2--12 & 10 & 47 \\
 2 & 1800 & ARG & 350 & 1--6 & 24 & 7 \\
 3 & 1726 & RES & 150 & 0--3 & 39 & 423 \\
 4 & 1772 & RES & 150 & 0--3 & 312 & 253 \\
 5 & 1805 & RES & 150 & 0--4 & 24 & 258 \\
 6 & 1800 & RES & 150 & 0--4 & 44 & 367 \\
 7 & 1569 & NAR & 250 & 0--30 & 0 & 0 \\
 8 & 723 & NAR & 650 & 0--60 & 0 & 1 \\
\hline
\end{tabular}
\end{table}

\subsubsection{Dataset} A number of different datasets have been used for English AES, including CLC-FCE \cite{yannakoudakis}, TOEFL11 \cite{toefl}, ICLE \cite{icle} and AAE \cite{aae}. However, these datasets are relatively small, with very few essays belonging to each prompt. As a result, we conduct our experiments on the Automated Student Assessment Prize (ASAP)\footnote{The dataset can be found at https://www.kaggle.com/c/asap-aes.} dataset, which is a large-scale, publicly-available dataset that has been widely adopted in the field of AES \cite{alikaniotis,taghipour,dong1,dong2,tay,phandi,cummins,jin}. This dataset comprises eight different essay sets, with essays in each set responding to a different prompt. The essays are written by students between Grade 7 and Grade 10. The dataset is summarised in Table \ref{dataset}.

\subsubsection{Cross Validation} Following Jin et al. \cite{jin}, we conduct eight-fold prompt-wise cross-validation, whereby for each fold, essays for one prompt are set aside for testing and the essays from the remaining prompts are used for training.

\subsubsection{Evaluation Metric} The scaled output score predictions are rescaled to their original scores. We then adopt the widely-used quadratic weighted kappa (QWK)\footnote{Refer to Appendix for details on how QWK is calculated.} \cite{taghipour,dong1,dong2,tay,phandi,jin} to evaluate the level of agreement between the human-rated scores and the predicted scores.

\subsubsection{Baseline Models}
We compare \textit{PAES} to the following baselines:
\begin{itemize}
    \item \textit{CNN-CNN-MoT:} This model \cite{dong1} represents essays as sequences of sentences and applies two CNN layers followed by a mean-over-time layer.
    \item \textit{CNN-LSTM-ATT:} This model \cite{dong2} employs a CNN followed by an attention pooling layer for sentence-level representations. It then uses an LSTM and an attention pooling layer to capture an essay-level representation.
    \item \textit{TDNN:} This two-stage approach \cite{jin} first uses RankSVM to assign pseudo labels. The second stage adopts a neural network featuring two layers of BiLSTMs on top of word embeddings and two layers of BiLSTM on top of a syntactic embedding, achieved through syntactic parsing.
    \item \textit{TDNN-mod:} Our reproduced results for \textit{TDNN} were considerably lower than published \cite{jin}, so we test a modified version \textit{TDNN-mod} whose two-stage approach applies the same principles as \textit{TDNN} while still achieving a good level of performance. \textit{TDNN-mod} extracts the same prompt-independent features as \textit{TDNN} and applies an SVM to assign pseudo labels to the lowest- and highest-quality essays in the target-prompt essay set. A neural network is then trained using the pseudo-labelled target-prompt essays as training data. This network comprises two \textit{CNN-LSTM-ATT}-based sub-networks over word embeddings and POS embeddings respectively. These two representations are concatenated together before the final score is predicted.
\end{itemize}
\textit{Note:} Results are not reported for the methods introduced by Cummins et al. \cite{cummins}, who only perform ranking rather than scoring when no target-prompt essays are available, or Phandi et al. \cite{phandi}, who report poor performance when only non target-prompt essays are available.

\subsection{Results}

\begin{table}
\caption{QWK scores for all models on ASAP dataset}\label{results}
\begin{tabular}{|p{3cm}|p{1cm}|p{1cm}|p{1cm}|p{1cm}|p{1cm}|p{1cm}|p{1cm}|p{1cm}|p{1cm}|}
\hline
& \multicolumn{8}{|c|}{\textbf{Prompts}} & \\
\hline
\textbf{Model} &  \textbf{1} & \textbf{2} & \textbf{3} & \textbf{4} & \textbf{5} & \textbf{6} & \textbf{7} & \textbf{8} & \textbf{Avg}\\
\hline
CNN-CNN-MoT & 0.281 & 0.500 & 0.105 & 0.275 & 0.408 & 0.312 & 0.236 & 0.170 & 0.286 \\
CNN-LSTM-ATT & 0.311 & 0.467 & 0.364 & 0.531 & 0.525 & 0.584 & 0.336 & 0.363 & 0.435 \\
TDNN* & 0.769 & \textbf{0.686} & 0.628 & \textbf{0.758} & 0.737 & \textbf{0.675} & 0.659 & 0.574 & \textbf{0.686} \\
TDNN-mod & 0.762 & 0.617 & 0.643 & 0.659 & \textbf{0.758} & 0.644 & 0.639 & 0.541 & 0.658 \\
PAES & \textbf{0.798} & 0.628 & \textbf{0.659} & 0.653 & 0.756 & 0.626 & \textbf{0.724} & \textbf{0.640} & \textbf{0.686} \\
\hline
\end{tabular}
\footnotesize{*Our results for \textit{TDNN} were considerably lower than those published by Jin et al. \cite{jin}, so we report the results published in their work.}
\end{table}

We list our results in Table \ref{results}. As can be seen, the models that only employ word embeddings as the essay representation (\textit{CNN-CNN-MoT} and \textit{CNN-LSTM-ATT}) lack robustness and produce poor results across the board. In contrast, the overall score for \textit{PAES} matches the state-of-the-art \textit{TDNN} and scores higher on five of the eight prompts. In the following section, our analyses show that \textit{PAES} is able to overcome the drawbacks of previous approaches.

\subsection{Analyses}
\subsubsection{Effect of Non Prompt-Specific Features}
The use of well-designed non prompt-specific features is important to the task of cross-prompt AES due to the need to represent essay quality in a general way. To demonstrate this, we conduct an experiment to investigate the addition of the features vector from \textit{PAES} to our baselines. In each case, the features vector is concatenated with the final essay representation before the score is predicted. The results are displayed in table \ref{featuresimp}.

\begin{table}
\caption{QWK scores for models with and without non prompt-specific features}\label{featuresimp}
\begin{tabular}{|p{3.5cm}|p{1cm}|p{1cm}|p{1cm}|p{1cm}|p{1cm}|p{1cm}|p{1cm}|p{1cm}|p{1cm}|}
\hline
& \multicolumn{8}{|c|}{\textbf{Prompts}} & \\
\hline
\textbf{Model} &  \textbf{1} & \textbf{2} & \textbf{3} & \textbf{4} & \textbf{5} & \textbf{6} & \textbf{7} & \textbf{8} & \textbf{Avg}\\
\hline
Features & 0.817 & 0.648 & 0.593 & 0.576 & 0.722 & 0.590 & 0.679 & 0.499 & 0.641 \\
\hline
CNN-CNN-MoT & 0.281 & 0.500 & 0.105 & 0.275 & 0.408 & 0.312 & 0.236 & 0.170 & 0.286 \\
CNN-CNN-MoT-Feat & \textbf{0.403} & \textbf{0.538} & \textbf{0.481} & \textbf{0.575} & \textbf{0.659} & \textbf{0.570} & \textbf{0.456} & \textbf{0.306} & \textbf{0.496} \\
\hline
CNN-LSTM-ATT & 0.311 & 0.467 & 0.364 & 0.531 & 0.525 & 0.584 & 0.336 & 0.363 & 0.435 \\
CNN-LSTM-ATT-Feat & \textbf{0.747} & \textbf{0.493} & \textbf{0.493} & \textbf{0.597} & \textbf{0.675} & \textbf{0.623} & \textbf{0.554} & \textbf{0.540} & \textbf{0.590} \\
\hline
\end{tabular}
\end{table}

From the results, we can see that the model \textit{Features}, which just takes the non prompt-specific features from \textit{PAES} as input to a linear layer and sigmoid activation, is a robust model achieving an average QWK score of 0.640. We also notice that for both baselines, concatenating the essay representation with the features vector increases performance significantly.

\subsubsection{Effect of POS Embedding}
We argue that directly using word embeddings to represent essays in the task of cross-prompt AES is unsuitable due to the issue of the model learning strong semantic representations and leading to overfitting the training data. To investigate this, we replace the word embeddings from the \textit{CNN-CNN-MoT} and \textit{CNN-LSTM-ATT} models with POS embeddings and compare the performance. The results are displayed in table \ref{posimp}. \par
\begin{table}
\caption{QWK scores for baseline models with and without POS embedding}\label{posimp}
\begin{tabular}{|p{3.5cm}|p{1cm}|p{1cm}|p{1cm}|p{1cm}|p{1cm}|p{1cm}|p{1cm}|p{1cm}|p{1cm}|}
\hline
& \multicolumn{8}{|c|}{\textbf{Prompts}} & \\
\hline
\textbf{Model} &  \textbf{1} & \textbf{2} & \textbf{3} & \textbf{4} & \textbf{5} & \textbf{6} & \textbf{7} & \textbf{8} & \textbf{Avg}\\
\hline
CNN-CNN-MoT & 0.281 & \textbf{0.500} & 0.105 & 0.275 & 0.408 & \textbf{0.312} & 0.236 & \textbf{0.170} & 0.286 \\
CNN-CNN-MoT-POS & \textbf{0.528} & 0.280 & \textbf{0.445} & \textbf{0.486} & \textbf{0.515} & 0.274 & \textbf{0.575} & 0.089 & \textbf{0.399} \\
\hline
CNN-LSTM-ATT & 0.311 & \textbf{0.467} & 0.364 & 0.531 & 0.525 & \textbf{0.584} & 0.336 & \textbf{0.363} & 0.435 \\
CNN-LSTM-ATT-POS & \textbf{0.324} & 0.324 & \textbf{0.586} & \textbf{0.612} & \textbf{0.741} & 0.541 & \textbf{0.520} & 0.272 & \textbf{0.490} \\
\hline
\end{tabular}
\end{table}
It can be seen from the results that using POS embeddings improves the performance of both models, with \textit{CNN-CNN-MoT} and \textit{CNN-LSTM-ATT} receiving overall performance improvements of 11.3\% and 5.5\% in terms of absolute value for the average QWK score across all prompts. This demonstrates that POS embeddings are more effective in representing essays in cross-prompt AES, due to the fact that they are able to achieve more generalized representations than those by word embeddings.

\subsubsection{Performance Robustness for Low-Resource Unlabelled Target-Prompt Essays} As we have stated, the \textit{TDNN} approach requires a sufficient quantity of unlabelled target-prompt essays in order to assign pseudo labels and train the second-stage neural network. To investigate this issue, we compare \textit{PAES} with \textit{TDNN-mod} for different sample sizes of target-prompt essays. As the reproduced \textit{TDNN} performs poorly, we compare our method with \textit{TDNN-mod}. Since \textit{TDNN-mod} is based on the same underlying principles as \textit{TDNN} -- they both utilize a large quantity of unlabelled target-prompt essays to assign pseudo labels that are used to train a neural network -- we make the reasonable assumption that the performance of both models will degrade in the same manner as fewer unlabelled target-prompt essays are made available. \par 
Testing is conducted on the full set of target-prompt essays, but only a percentage of those essays is available for training. Since \textit{PAES} doesn't use target-prompt essays in the training phase, the only difference is that the features for the test set are normalized based on the values from the available essays. The results of our experiments are shown in Fig. 2.

\begin{tikzpicture}
\begin{axis}
[
title={\textbf{Fig. 2.} QWK scores for different quantities of target-prompt essays},
xlabel={\% of target-prompt essays available}, ylabel={Overall QWK Score}, xmin=0, xmax=50,
ymin=0.5, ymax=0.7,
height=5.5cm,
width=11.5cm,
legend columns=5,
legend pos=south east,
ticklabel style = {font=\large},
label style = {font=\large},
xtick={10, 20, 30, 40},
ytick={0.5,0.55,0.6,0.65,0.7},
]

\addplot
[
line width=1pt,
color=green,
mark=square,
]
coordinates{
    (10,0.556) (20,0.608) (30,0.612) (40,0.648)};

\addplot
[
line width=1pt,
color=RedViolet,
mark=halfsquare*,
]
coordinates{
    (10,0.650) (20,0.675) (30,0.677) (40,0.678)};

\legend{$TDNN-mod$,$PAES$}
\end{axis}
\end{tikzpicture}

As can be seen from Fig. 2, \textit{TDNN-mod} still performs well when 40\% of the unlabelled target-prompt essays are available as there are still enough essays to use as training data for the neural network in the second stage. However, as the number of available target-prompt essays decreases, the performance of \textit{TDNN-mod} decreases, dropping down to 0.556 when 10\% of the target-prompt essays are available. In contrast, since \textit{PAES} doesn't use the target-prompt essays in its training phase, its performance remains strong, with only a slight decrease in performance at 10\% due to our using the available target-prompt essays to normalize feature values of the entire test set. The consistency of our model in this experiment demonstrates that our model is effective and robust, even in the event of there being a low quantity of unlabelled target-prompt essays available.

\subsubsection{Effect of score distributions in target-prompt set} The performance of \textit{TDNN} \cite{jin} and \textit{TDNN-mod} are dependent on the score distributions of the target-prompt essay set. If few essays are of extreme good or bad quality, then a number of pseudo labels will have a high error with regards to the gold score. To examine this intuition, we analyse the number of minimum and maximum scores from each essay set, displayed in the last two columns of Table \ref{dataset}. \par
As can be seen, none of the essays in Set 7 were awarded either the minimum score of 0 or the maximum score of 30. Also, for Set 8, none were awarded the minimum score of 0, while only one was awarded the maximum score of 60. A closer inspection of the data reveals that only two essays in Set 8 were given a score of 15 or lower. Here, it is evident that the pseudo-labelling method from the two-stage approaches will be less effective in these cases because the pseudo labels will be a long way from the gold values. This is evident in the results in Table \ref{results}, where the performance of both \textit{TDNN} and \textit{TDNN-mod} on Sets 7 and 8 is considerably lower than \textit{PAES}, which doesn't use pseudo labelling. In contrast, we can also observe that a high number of essays with extreme quality are present in sets 3--6. This seems to have a positive impact on the performance of the two-stage models, which both score highly on these prompts.

\section{Conclusion} \label{conclusions}
This paper aims to address the issues present in cross-prompt AES, in which no target-prompt essays are available for training. Through thorough experimentation and analysis, we demonstrate the shortcomings of leading prompt-specific AES approaches \cite{taghipour,dong1,dong2} and the leading cross-prompt method \cite{jin} in the task of cross-prompt AES. Our \textit{PAES} method is the first to use a neural-network combined with non prompt-specific features in a single-stage approach for cross-prompt AES. We utilize POS embeddings to achieve syntactic representations, avoiding the issue of overfitting to the training data, as with the prompt-specific approaches. We also extract a set of features that capture essay quality from a variety of perspectives to achieve a good general representation of essay quality. \par
Our single-stage approach avoids the need to use unlabelled target-prompt essays in the training phase, thus eliminating the issues of insufficient unlabelled target-prompt essays and target-prompt essay-quality distribution that exist in \textit{TDNN} \cite{jin}. Experiments carried out on the widely-used ASAP dataset show that our model is robust and is able to achieve state-of-the-art performance.

%
%
%

\begin{thebibliography}{8}

\bibitem{alikaniotis}
Alikaniotis, D., Yannakoudakis, H., Rei, M.: Automatic text scoring using neural networks. In Proc. of ACL, pages 715--725 (2016)

\bibitem{attali}
Attali, Y., Burstein, J.: Automated essay scoring with e-rater\textcircled{R} v. 2. The Journal of Technology,
Learning and Assessment 4(3) (2006)

\bibitem{toefl}Blanchard, D., Tetreault, J.,
Higgins, D., Cahill, A., Chodorow, M.
TOEFL11: A corpus of non-native English. ETS Research
Report Series, 2013(2):i--15 (2013)

\bibitem{chen}
Chen, H., He, B.: Automated essay scoring by maximizing human-machine agreement. In Proceedings of the 2013 Conference on Empirical Methods in Natural Language Processing (2013)

\bibitem{cummins}
Cummins, R., Zhang, M., Briscoe, T.: Constrained multi-task learning for automated essay scoring. In Proc. of ACL, pages 789--799 (2016)

\bibitem{dauphin}
Dauphin, Y., de Vries, H., Bengio, Y.: Equilibrated adaptive learning rates for nonconvex optimization. In Advances in Neural Information Processing Systems. pages 1504--1512 (2015)

\bibitem{dong1} 
Dong, F., Zhang, Y.: Automatic features for essay scoring -- An empirical study. In Proc. of EMNLP, pages 1072--1077 (2016)

\bibitem{dong2}
Dong, F., Zhang, Y., Yang, J.: Attention-based recurrent convolutional neural network for automatic essay scoring. In Proc. of CoNLL, pages 153--162, (2017)

\bibitem{icle}
Granger, S., Dagneaux, E., Meunier, F., Paquot, M. International Corpus
of Learner English (Version 2). Presses universitaires de
Louvain (2009)

\bibitem{hochreiter}
Hochreiter, S., Schmidhuber, J.: Long short-term memory. Neural computation 9(8):1735--1780 (1997)

\bibitem{jin}
Jin, C., He, B., Hui, K., Sun, L.: TDNN: A two-stage deep neural network for promptindependent automated essay scoring. In Proc. of ACL, pages 1088--1097 (2018)

\bibitem{ke}
Ke, Z., Ng, V.: Automated essay scoring: A survey of the state of the art. In Proc. of the 28th IJCAI, pages 6300--6308. International Joint Conferences on Artificial Intelligence Organization (2019)

\bibitem{larkey}
Larkey, L., S.: Automatic essay grading using text categorization techniques. In SIGIR. ACM, pages 90--95 (1998)

\bibitem{lei}
Lei, C., Man, K., L., Ting, T.: Using learning analytics to analyze writing skills of students: A case study in a technological common core curriculum course. IAENG International Journal of Computer Science, 41(3) (2014)

\bibitem{mcnamara}
Mcnamara, D., S., Crossley, S., A., Roscoe, R., D., Allen, L., K., Dai, J.: A hierarchical classification approach to automated essay scoring. Assessing Writing 23:35--59 (2015)

\bibitem{page}
Page, E., B.: The imminence of... grading essays by computer. The Phi Delta Kappan, 47(5):238--243 (1966)

\bibitem{pascanu}
Pascanu, R., Mikolov, T., Bengio, Y.: On the difficulty of training recurrent neural networks. ICML (3) 28:1310--1318 (2013)

\bibitem{phandi}
Phandi, P., Chai, K., M., A., Ng, H., T.: Flexible domain adaptation for automated essay scoring using correlated linear regression. In Proc. of EMNLP, pages 431--439 (2015)

\bibitem{rudner}
Rudner, L., M.: Automated essay scoring using bayes’ theorem. National Council on Measurement in Education New Orleans La 1(2):3--21 (2002)

\bibitem{smith}
Smith, M., Taffler, R.: Readability and understandability: Different measures of the textual complexity of accounting narrative. Accounting, Auditing \& Accountability Journal, 5(4):0--0 (1992)

\bibitem{aae}
Stab, C., Gurevych, I. Annotating argument components and relations in persuasive essays. In Proc. of COLING, pages
1501--1510, (2014)

\bibitem{sutskever}
Sutskever, I., Vinyals, O., Le, Q., V.: Sequence to sequence learning with neural networks. In Proc. of NIPS, pages 3104--3112 (2014)

\bibitem{taghipour}
Taghipour, K., Ng, H., T.: A neural approach to automated essay scoring. In Proc. of EMNLP, pages 1882--1891, (2016)

\bibitem{tay}
Tay, Y., Phan, M., C., Tuan, L., A., Hui, S., C.: SkipFlow: Incorporating neural coherence features for end-to-end automatic text scoring. In Proc. of AAAI, pages 5948--5955 (2018)

\bibitem{williamson}
Williamson, D., M.: A framework for implementing automated scoring. In Annual Meeting of the American Educational Research Association and the National Council on Measurement in Education, San Diego, CA. (2009)

\bibitem{yannakoudakis}
Yannakoudakis, H., Briscoe, T., Medlock, B.: A new dataset and method for automatically grading ESOL texts. In Proc. of ACL, pages 180--189 (2011)

\end{thebibliography}
%

\newpage
\section*{Appendix}

\subsection*{PAES Model Components}
\subsubsection{Convolutional Layer} Following Dong et al. \cite{dong2}, we apply a 1D convolutional layer followed by attention pooling to capture sentence-level representations. The convolutional layer is applied to each sentence:
\begin{equation} \label{eq2}
\textbf{z}_i = f(\textbf{W}_z \cdot[\textbf{x}_i:\textbf{x}_{i+h_w-1}]+b_z)
\end{equation}

where \(\textbf{W}_z\) is the weights matrix, \(b_z\) is the bias vector, and \(h_w\) is the size of the convolution window. \par
The attention pooling layer applied to the output of the convolutional layer is designed to capture the sentence representations and is defined as follows:
\begin{equation} \label{eq3}
\textbf{m}_i = tanh(\textbf{W}_m \cdot \textbf{z}_i + \textbf{b}_m)
\end{equation}
\begin{equation} \label{eq4}
u_i = \frac{e^{\textbf{w}_u \cdot \textbf{m}_i}}{\sum e^{\textbf{w}_u \cdot \textbf{m}_j}}
\end{equation}
\begin{equation} \label{eq5}
\textbf{s} = \sum u_i \textbf{z}_i
\end{equation}

where \(\textbf{W}_m\) is a weights matrix, \(\textbf{w}_u\) is a weights vector, \(\textbf{m}_i\) is the attention vector for the \(i\)-th word, \(u_i\) is the attention weight for the \(i\)-th word, and \(\textbf{s}\) is the final sentence representation.

\subsubsection{Recurrent Layer} As with Dong et al. \cite{dong2}, we apply an LSTM layer \cite{hochreiter,pascanu} followed by attention pooling. From the output of the previous layer, the representation consists of \(T\) sentences, \(s_1,s_2,...,s_T\). The LSTM is applied to each sentence, whereby the output \(\textbf{h}_t\) is the output at time \(t\) given the sentence input \(s_t\) and output of previous timestep \(\textbf{h}_{t-1}\):

\begin{equation} \label{eq6}
\textbf{i}_t = \sigma(\textbf{W}_i\cdot\textbf{s}_t + \textbf{U}_i \cdot \textbf{h}_{t-1} + \textbf{b}_i)
\end{equation}
\begin{equation} \label{eq7}
\textbf{f}_t = \sigma(\textbf{W}_f\cdot\textbf{s}_t + \textbf{U}_f \cdot \textbf{h}_{t-1} + \textbf{b}_f)
\end{equation}
\begin{equation} \label{eq8}
\textbf{$\tilde{c}$}_t = tanh(\textbf{W}_c \cdot \textbf{s}_t + \textbf{U}_c \cdot \textbf{h}_{t-1} + \textbf{b}_c)
\end{equation}
\begin{equation} \label{eq9}
\textbf{c}_t = \textbf{i}_t \circ \textbf{$\tilde{c}$}_t + \textbf{f}_t \circ \textbf{c}_{t-1}
\end{equation}
\begin{equation} \label{eq10}
\textbf{o}_t = \sigma(\textbf{W}_o\cdot\textbf{s}_t + \textbf{U}_o \cdot \textbf{h}_{t-1} + \textbf{b}_o)
\end{equation}
\begin{equation} \label{eq11}
\textbf{h}_t = \textbf{o}_t \circ tanh(\textbf{c}_t)
\end{equation}

where \(\textbf{s}_t\) and \(\textbf{h}_t\) are the input sentence and output state at time \(t\). \(\textbf{W}_i\), \(\textbf{W}_f\), \(\textbf{W}_c\), \(\textbf{W}_o\), \(\textbf{U}_i\), \(\textbf{U}_f\), \(\textbf{W}_c\) and \(\textbf{U}_o\) are weights matrices; \(\textbf{b}_i\), \(\textbf{b}_f\), \(\textbf{b}_c\) and \(\textbf{b}_o\) are bias vectors; and \(\sigma\) denotes the sigmoid activation function. \par
To obtain the essay representation, attention pooling is applied to the outputs \(\textbf{h}_1,\textbf{h}_2,...,\textbf{h}_T \) from the recurrent layer:

\begin{equation} \label{eq12}
\textbf{a}_i = tanh(\textbf{W}_a \cdot \textbf{h}_i + \textbf{b}_a)
\end{equation}
\begin{equation} \label{eq13}
\alpha_i = \frac{e^{\textbf{w}_\alpha \cdot \textbf{a}_i}}{\sum e^{\textbf{w}_\alpha \cdot \textbf{a}_j}}
\end{equation}
\begin{equation} \label{eq14}
\textbf{o} = \sum \alpha_i \textbf{h}_i
\end{equation}

where \(\textbf{W}_a\) is a weights matrix, \(\textbf{w}_\alpha\) is a weights vector, \(\textbf{a}_i\) is the attention vector for the \(i\)-th sentence, \(\alpha_i\) is the attention weight for the \(i\)-th sentence, and \(\textbf{o}\) is the overall essay representation.

\subsection*{Training} \label{training}
\begin{table}
\begin{center}
    \caption{\textit{PAES} component hyper-parameters}\label{paesparams}
    \begin{tabular}{|p{4.5cm}|p{4cm}|p{1cm}|}
    \hline
    \textbf{Component} & \textbf{Hyper-parameter} & \textbf{Value} \\
    \hline
     POS Embeddings & Output Vector Dimensions & 50 \\
     \hline
     \multirow{2}{7em}{Convolutional Layer} & Number of Filters & 100 \\
     & Filter Length & 5 \\ 
     \hline
     LSTM Layer & Output Dimensions & 100 \\
     \hline
     Dropout & Probability & 0.5 \\
     \hline
     Non Prompt-specific Features & Vector Dimensions & 86 \\
    \hline
    \end{tabular}
\end{center}
\end{table}

\subsubsection{Objective} We adopt the mean square error (MSE) objective function over which to optimize our model. Given that there are \(N\) essays, MSE calculates the average square difference between the predicted score \(\hat{y}_i\) and the gold score \(y_i\). Formally, it is calculated as follows:

\begin{equation} \label{eq17}
MSE(y, \hat{y}) = \frac{1}{N} \sum_{i=1}^N (\hat{y_i} - y_i)^2
\end{equation}

\subsubsection{Optimizer} We use the RMSprop \cite{dauphin} algorithm to optimize our model with the learning rate \(\eta\) set to 0.001. We also apply dropout with a drop rate probability of 0.5 to avoid the issue of overfitting.

\subsection*{Quadratic Weighted Kappa}
Quadratic weighted kappa (QWK) is defined as follows:

\begin{equation} \label{eq18}
W_{i,j} = \frac{(i-j)^2}{(R-1)^2}
\end{equation}

where \(i\) and \(j\) are the human score and predicted score, respectively. \(R\) is the number of possible ratings. \par
An \(R\)-by-\(R\) dimension matrix \(O\) is created, where the term \(O_{i,j}\) records the number of occurrences of a score given by human rater \(i\) and prediction \(j\). An expected score matrix \(E\) with \(R\)-by-\(R\) dimensions is also calculated. It is calculated as the outer product between the human rater and prediction histogram vector. The values of \(E\) and \(O\) are normalized so that the two matrices have the same sum. The QWK value is then calculated as follows:

\begin{equation} \label{eq19}
K = 1 - \frac{\sum W_{i,j}O_{i,j}}{\sum W_{i,j}E_{i,j}}
\end{equation}

\end{document}